\documentclass[runningheads]{llncs}
\usepackage{enumitem}
\usepackage[T1]{fontenc}
\usepackage[utf8]{inputenc}
\usepackage{graphicx}
\usepackage{booktabs}
\usepackage{tabularx}
\usepackage{makecell}
\usepackage{multirow}
\usepackage{array}
\usepackage{amssymb}
\usepackage[misc]{ifsym}
\usepackage{placeins}
\usepackage{xcolor}
\newcommand{\corr}{(\Letter)}
\usepackage{mwe}

\usepackage{float}   
\usepackage{hyperref}
\usepackage{amsmath}
\usepackage{graphicx}
\usepackage{comment}
\usepackage{wrapfig}
\usepackage{kotex}
\usepackage{todonotes}
\usepackage{siunitx}
\usepackage[table]{xcolor}
\usepackage{booktabs}
\usepackage{multirow}
\definecolor{oursbg}{RGB}{220,232,246}

\definecolor{darkgreen}{RGB}{0,120,0}
\definecolor{darkred}{RGB}{180,40,40}
\newcommand{\improve}[1]{\textcolor{darkgreen}{\ensuremath{#1}}}
\newcommand{\worse}[1]{\textcolor{darkred}{\ensuremath{#1}}}
\newcommand{\same}[1]{\ensuremath{#1}}

\newcommand{\bftab}[1]{{\fontseries{b}\selectfont #1}}

\usepackage{tikz}
\usepackage{pgfplots}
\pgfplotsset{compat=1.18}

\presetkeys{todonotes}{inline, color=topaz!40, bordercolor=topaz}{}

\begin{document}

\title{Modeling Matches as Language: \\ A Generative Transformer Approach for Counterfactual Player Valuation in Football}
\toctitle{Modeling Matches as Language: A Generative Transformer Approach for Counterfactual Player Valuation in Football}

\titlerunning{A Generative Transformer Approach for Player Valuation in Football}

\author{
Miru Hong\inst{1} \and
Minho Lee\inst{2} \and
Geonhee Jo\inst{1} \and
Hyeokje Cho\inst{1} \and \\
Hyunsung Kim\inst{3,4} \and
Pascal Bauer\inst{2,5} \and
Sang-Ki Ko\inst{1}\corr
}

\authorrunning{M. Hong et al.}
\tocauthor{Miru Hong, Minho Lee, Geonhee Jo, Hyeokje Cho, Hyunsung Kim, Pascal Bauer, Sang-Ki Ko}

\institute{
University of Seoul, Seoul, Republic of Korea \\
\email{\{mirunoyume,geonhee,brandon56,sangkiko\}@uos.ac.kr}
\and
Saarland University, Saarbrücken, Germany \\
\email{\{minho.lee,pascal.bauer\}@uni-saarland.de}
\and
KAIST, Daejeon, Republic of Korea \\
\email{hyunsung.kim@kaist.ac.kr}
\and
Fitogether Inc., Seoul, Republic of Korea
\and
Deutscher Fu\ss{}ball-Bund, Frankfurt, Germany
}

\maketitle              

\begin{abstract}
Evaluating football player transfers is challenging because player actions depend strongly on tactical systems, teammates, and match context.
Despite this complexity, recruitment decisions often rely on static statistics and subjective expert judgment, which do not fully account for these contextual factors. This limitation stems largely from the absence of counterfactual simulation mechanisms capable of predicting outcomes in hypothetical scenarios. To address these challenges, we propose ScoutGPT, a generative model that treats football match events as sequential tokens within a language modeling framework. Utilizing a NanoGPT-based Transformer architecture trained on next-token prediction, ScoutGPT learns the dynamics of match event sequences to simulate event sequences under hypothetical lineups, demonstrating superior predictive performance compared to existing baseline models. Leveraging this capability, the model employs Monte Carlo sampling to enable counterfactual simulation, allowing for the assessment of unobserved scenarios. Experiments on K League data show that simulated player transfers lead to measurable changes in offensive progression and goal probabilities, indicating that ScoutGPT captures player-specific impact beyond traditional static metrics.

\end{abstract}

\section{Introduction}

Evaluating individual contribution is challenging in complex multi-agent environments, where behavior depends not only on an agent’s own ability but also on interactions with surrounding agents and context. Football provides a particularly demanding instance of this problem: player actions are shaped by tactical roles, teammates, opponents, and match state. As a result, player transfer evaluation cannot be reduced to a like-for-like replacement problem, since moving a player to a new team alters the tactical configuration and reshapes interaction patterns on the pitch. Transfer evaluation therefore requires estimating how a player will behave under this distribution shift, rather than extrapolating directly from past performance alone.

Previous approaches only partially address this problem. Event-based evaluation
frameworks~\cite{Bransen2020Player,Decroos2019Actions,Liu2020Deep,Luo2020Inverse,Pappalardo2019PlayeRank,Singh2019Introducing} quantify observed events but do not generate how action sequences would evolve under a new tactical context, and projection systems in other sports operate on aggregate season outcomes that miss on-pitch micro-interactions. Recent generative models focus either on continuous 
trajectories~\cite{Capellera2024TranSPORTmer,Capellera2026JointDiff,Capellera2025Unified,Fassmeyer2022Semi,Xu2023Uncovering,Xu2025SportsTraj}, which capture spatial movement but not the tactical semantics of discrete events, or on next-event
prediction~\cite{MendesNeves2024Forecasting,MendesNeves2024Towards,MendesNeves2026Scalable,Simpson2022Seq2Event,Yeung2025Transformer}, which
targets observed continuations rather than hypothetical-transfer sequences;
even event-level estimators such as On-Ball Value~\cite{Hong2025EventGPT}
generate only short fragments of play. Evaluating transfer scenarios instead
requires generating full event sequences under a new lineup and match context,
enabling value computation over the entire simulated possession.



To address this problem, we introduce ScoutGPT, an autoregressive generative framework for football event streams related to Large Event Models (LEMs)~\cite{MendesNeves2026Scalable}. ScoutGPT treats a match as a structured sequence in which each event is decomposed into discrete attributes through tokenization and predicted sequentially via next-token prediction, conditioned on player identity and match context. Alongside next-action prediction, the model estimates goal-scoring and goal-conceding probabilities at each step, aligning generated sequences with match value (VAEP)~\cite{Decroos2019Actions} and supporting event-level simulation of hypothetical player transfers under new tactical environments~\cite{Dinsdale2022Transfer,VanArem2025Forecasting}.

    
    
To summarize, our main contributions are as follows:
\begin{itemize}[leftmargin=*, topsep=0pt, itemsep=1pt, parsep=0pt, partopsep=0pt]
    \item \textbf{Structured Event Modeling for Context-Aware Simulation:} We introduce a fine-grained tokenization scheme that decomposes football events into semantic components (e.g., actor, location, and action type). This structure enables ScoutGPT to capture dependencies across event attributes and model football event sequences at a finer granularity.
    
    \item \textbf{Value-Aware Generative Modeling:} We propose a multi-task learning objective that combines next-token prediction with explicit scoring and conceding probability estimation. This design encourages the model to reflect both event likelihood and match value, and improves predictive performance over non-value-aware variants.
    
    \item \textbf{Counterfactual Simulation for Player Recruitment:} We show that ScoutGPT can simulate how a player's on-ball contribution profile shifts in a new tactical environment, supporting data-driven analysis of transfer fit.
\end{itemize}
    
\begin{figure}[t!]
    \centering
    \includegraphics[width=\linewidth]{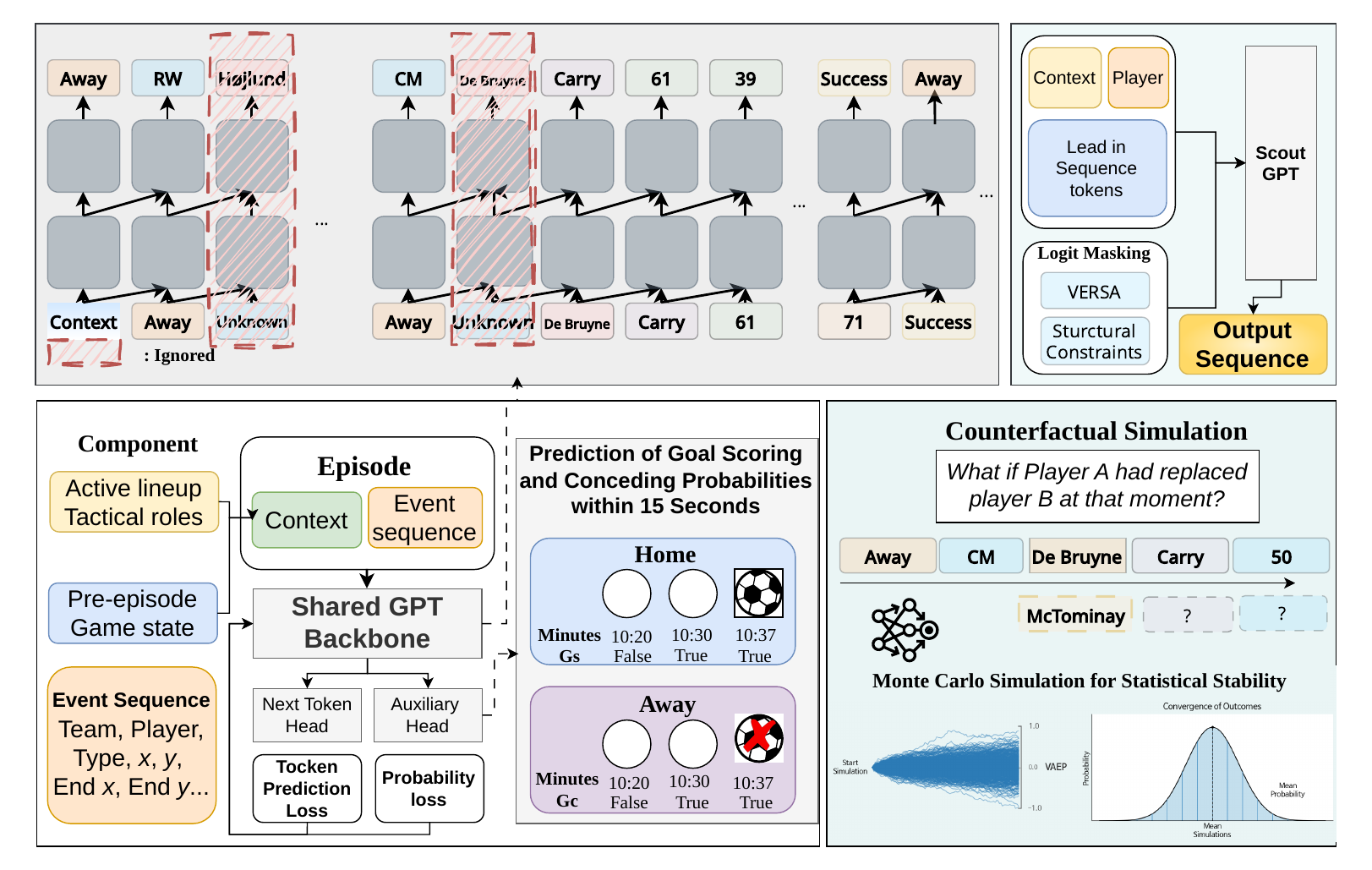}
    \caption{Overview of the ScoutGPT framework. Our NanoGPT-based Transformer model autoregressively predicts event tokens, enabling counterfactual `what-if' simulations. For instance, replacing Kevin De Bruyne with Scott McTominay could alter actions (e.g., pass/shot) or modify the same action with a different location, outcome, or VAEP.
    }
    \label{fig:overview}
\end{figure}

\section{Related Work}
Our work sits at the intersection of three lines of research: data-driven player valuation, generative modeling of sports event streams, and counterfactual simulation for player transfers.

\subsection{Data-Driven Player Valuation}
Action-value frameworks have become the standard for data-driven player valuation. VAEP quantifies player contribution by aggregating short-horizon changes in scoring and conceding probabilities across all on-ball actions~\cite{Decroos2019Actions}, while EPV decomposes instantaneous possession value into interpretable subcomponents~\cite{Fernandez2019Decomposing,Fernandez2021Framework}. PlayeRank extends this further by constructing multi-dimensional, role-aware player ratings from large-scale event logs~\cite{Pappalardo2019PlayeRank}. Collectively, these methods provide strong discriminative estimators for observed behavior. However, they evaluate actions that have already occurred and are not designed to generate counterfactual event sequences under hypothetical team configurations---a requirement that arises when assessing transfer fit.

\subsection{Generative Sequence Modeling of Sports Events}
A growing body of work frames football events as structured sequential prediction
problems. Seq2Event~\cite{Simpson2022Seq2Event} and Large Event Models
(LEMs)~\cite{MendesNeves2026Scalable} decompose each event into multiple
attributes and roll out match continuations from a given state, while
NMSTPP~\cite{Yeung2025Transformer} and related neural point process
models~\cite{Du2016Recurrent,Zuo2020Transformer} extend this to continuous-time
streams with explicit timing and mark distributions. Transformer
architectures~\cite{Vaswani2017Attention,Brown2020Large} have likewise been
adapted to predict matches as autoregressive token
sequences~\cite{Adjileye2024RisingBALLER,Baron2024Foundation,MendesNeves2024Forecasting,MendesNeves2024Towards}.
Despite strong short-horizon accuracy, these approaches optimize primarily for sequence likelihood without goal-oriented supervision, so they do
not account for the tactical value of decisions. Two further gaps limit their use for counterfactual transfer simulation. First, entity-conditioning
for player substitution is absent or indirect, making it difficult to hold context fixed while replacing a specific player. Second, unconstrained
generation can produce logically inconsistent transitions over longer horizons. ScoutGPT addresses these limitations by pairing the autoregressive
objective with explicit value supervision and VERSA-based constraint masking~\cite{Jo2026VERSA}.

\subsection{Counterfactual Simulation in Sports}
Macro-level transfer forecasting---baseball projection systems (ZiPS, PECOTA) and
football ability-curve regression~\cite{VanArem2025Forecasting}---predicts
aggregate season statistics from historical and age-curve data, too coarse to
capture event-level tactical dynamics. Graph-based methods recommend
positionally similar replacements from a relational player
network~\cite{Yilmaz2022Learning} but do not model how behavior changes in a new
team context, while hierarchical Bayesian xG~\cite{Mahmudlu2025What} and causal
evaluation frameworks~\cite{Susmann2026Counterfactual} isolate the counterfactual
impact of individual actions yet cannot generate the event sequences needed to
assess a full transfer. Closest to our setting, TacEleven~\cite{Zhao2025TacEleven}
uses language models to explore attacking tactics but only over fragmented paths,
and EventGPT~\cite{Hong2025EventGPT} generates only short fragments, approximating
the remaining value via residual OBV rather than from fully simulated sequences.

\section{Methodology}
\label{sec:method}

This section presents the pipeline of ScoutGPT. Section~\ref{subsec:problem} formulates football event modeling as sequence prediction. Section~\ref{subsec:tokenization} describes how the match context and events are serialized into a single token sequence. Section~\ref{subsec:architecture} then details the core architecture consisting of a backbone and two task-specific prediction heads, and Section~\ref{subsec:objective} introduces the objective that jointly train the model for both tasks. Finally, Section~\ref{subsec:inference} describes the constrained decoding procedure used to generate logically valid event sequences at inference time.



\subsection{Problem Formulation}
\label{subsec:problem}

We formulate football event modeling as a sequence prediction task conditioned on a global match context. Let $\mathbf{c}$ denote a 54-dimensional match context vector and $(\mathbf{e}_1, \mathbf{e}_2, \ldots, \mathbf{e}_T)$ denote a sequence of on-ball events occurring under this context, where each event $\mathbf{e}_t$ is a 10-dimensional tuple of integer tokens (See Table~\ref{tab:input_schema} describing the components of $\mathbf{c}$ and $\mathbf{e}_t$). Given the context $\mathbf{c}$ and the past events $\mathbf{e}_{1:t}$, our objective is to jointly predict (i) the next event $\mathbf{e}_{t+1}$ and (ii) two binary goal indicators $g^+_t, g^-_t \in \{0, 1\}$ denoting whether the acting team scores ($g^+_t = 1$) or concedes ($g^-_t = 1$) a goal within 15 seconds after $\mathbf{e}_{t}$.
Equivalently, the model estimates the next-event distribution $P(\mathbf{e}_{t+1} \mid \mathbf{c}, \mathbf{e}_{1:t}; \theta)$ together with the goal probabilities $P(g^+_t = 1 \mid \mathbf{c}, \mathbf{e}_{1:t}; \theta)$ and $P(g^-_t = 1 \mid \mathbf{c}, \mathbf{e}_{1:t}; \theta)$, where $\theta$ denotes the model parameters.

\begin{table}[b!]
\centering
\setlength{\tabcolsep}{4pt}
\footnotesize
\caption{Components of the context block $\mathbf{c}$ and each event tuple $\mathbf{e}_t$. Note that continuous fields (locations and elapsed time) are quantized into discrete bins.}
\label{tab:input_schema}
\begin{tabularx}{\textwidth}{@{}llXc@{}}
\toprule
Group & Symbol & Description & \# Tokens \\
\midrule
\multirow{5}{*}{\makecell{Context \\ $\mathbf{c}$}}
& $u_{\text{H}}, u_{\text{A}}$ & Team IDs of the home and away teams & $2$ \\
& $r_k$ & Role$^{\dagger}$ of each player $k$ & $22$ \\
& $p_k$ & Player ID of each player $k$ & $22$ \\
& $\mathbf{q}$ & Match state: period, minutes, home/away goals, home/away yellow and red cards & $8$ \\
\midrule
\multirow{8}{*}{\makecell{Event \\ $\mathbf{e}_t$}}
& $u_t$ & Team ID of the acting team & $1$ \\
& $r_t$ & Role$^{\dagger}$ of the acting player & $1$ \\
& $p_t$ & Player ID of the acting player & $1$ \\
& $a_t$ & Action type$^{\dagger}$ (e.g., Pass, Carry, Cross, Shot) & $1$ \\
& $x^{\rm start}_t, y^{\rm start}_t$ & Start location in meters, $\{0, \ldots, 105\} \times \{0, \ldots, 68\}$ & $2$ \\
& $x^{\rm end}_t, y^{\rm end}_t$ & End location in meters, $\{0, \ldots, 105\} \times \{0, \ldots, 68\}$ & $2$ \\
& $\delta_t$ & Time elapsed in seconds since the previous event & $1$ \\
& $o_t$ & Action outcome (success or failure) & $1$ \\
\bottomrule
\end{tabularx}
{\footnotesize $^{\dagger}$ The full lists of roles and action types are provided in the Supplementary Material.}
\end{table}

\subsection{Structured Event Tokenization}
\label{subsec:tokenization}


The match context $\mathbf{c}$ and the events $\mathbf{e}_{1:T}$ contain both categorical attributes and quantized continuous ones, together forming a heterogeneous input to the model. To process this hybrid data structure with a Transformer~\cite{Vaswani2017Attention}, we adopt a tokenization strategy that flattens both $\mathbf{c}$ and $\mathbf{e}_{1:T}$ into a single token sequence. The context $\mathbf{c}$ is laid out as its 54 constituent tokens $(c_1, \dots, c_{54})$, and each event $\mathbf{e}_t$ is unfolded into its 10 atomic tokens $(s_{t,1}, \dots, s_{t,10})$. 
The full input sequence $\mathbf{s}$ is then formed by placing the context tokens first, followed by the flattened event tokens in temporal order.
\begin{equation}
    \mathbf{s} = (\underbrace{c_1, \dots, c_{54}}_{\mathbf{c}}, \underbrace{s_{1,1}, \dots, s_{1,10}}_{\mathbf{e}_1}, \dots, \underbrace{s_{T,1}, \dots, s_{T,10}}_{\mathbf{e}_T}).
    \label{eq:flatten}
\end{equation}

Since meaningful dependencies arise primarily among events that occur consecutively while the ball is in play, we partition each match into a set of in-play segments called \emph{episodes}~\cite{Kim2023Ball} and perform sequence prediction independently within each episode. Each episode corresponds to a single in-play segment, starting with a set-piece or a kick off
and ending with a goal, a foul, or the ball going out of play, so it corresponds to a single coherent phase of play rather than an arbitrary time window. Since episodes have varying lengths, we cap the input sequence at $T_\text{max}=100$ events and split longer episodes into overlapping chunks via a sliding window with stride $T_\text{max}/2=50$.
The match state $\mathbf{q}$ corresponds to the moment that each episode begins, so the 54 context tokens are shared across all chunks in the episode, and only the event-token window slides forward.



\subsection{Model Architecture}
\label{subsec:architecture}


Through the tokenization in Section~\ref{subsec:tokenization}, the next-event prediction in Section~\ref{subsec:problem} reduces to next-token prediction over the flattened sequence $\mathbf{s}$. That is, conditioned on the context block and the preceding event tokens, ScoutGPT predicts the next event token of $\mathbf{s}$ one at a time, generating each event field by field. At the last token of each event, it additionally predicts the goal-scoring and goal-conceding indicators defined in Section~\ref{subsec:problem}. The model utilizes the NanoGPT architecture\footnote{\url{https://github.com/karpathy/nanoGPT}}, an efficient implementation of the GPT-2 decoder-only Transformer~\cite{Radford2019Language}, comprising a shared backbone and task-specific prediction heads.


\subsubsection{Backbone}
Given an input sequence $\mathbf{s} \in \mathbb{Z}^N$ of length $N$, the model maps each integer token to a dense vector by indexing into a learnable token-embedding matrix $W_{\rm tok} \in \mathbb{R}^{|\mathcal{V}| \times d}$ and adds learnable positional embeddings $W_{\rm pos} \in \mathbb{R}^{N_{\rm max} \times d}$ to obtain the initial hidden states:
\begin{equation}
    \mathbf{h}^{(0)} = W_{\rm tok}[\mathbf{s}] + W_{\rm pos}[{:}N] \in \mathbb{R}^{N \times d},
\end{equation}
where $\mathcal{V}$ is the shared token-embedding vocabulary, $N_{\rm max}$ is the maximum sequence length in tokens, and $d$ is the embedding dimension. Note that $\mathcal{V}$ is the union of sub-vocabularies of all fields, so $W_{\rm tok}$ stores the embeddings of every field in one shared table.

The two field types are organized differently within this table. Each categorical field (i.e., team ID, role, player ID, action type, and action outcome in Table~\ref{tab:input_schema}) owns a separate sub-vocabulary, and thus a disjoint block of rows in $W_{\rm tok}$, so that its category labels are embedded independently of the other fields. The quantized continuous fields (i.e., the start/end locations and the elapsed time) instead share a single sub-vocabulary of integer bins spanning 0 to 105, so a given bin index maps to the same row of $W_{\rm tok}$, regardless of which continuous field it corresponds to.



Once the initial hidden states $\mathbf{h}^{(0)}$ are obtained, they are processed by a stack of $L$ Pre-LayerNorm Transformer blocks~\cite{Xiong2020LayerNorm}.
Each block of layer $l = 0, \ldots, L-1$ takes $\mathbf{h}^{(l)}$ as input and computes $\mathbf{h}^{(l+1)}$ as follows:
\begin{equation}
    \tilde{\mathbf{h}}^{(l)} = \text{MSA}(\text{LN}(\mathbf{h}^{(l)})) + \mathbf{h}^{(l)}, \quad \mathbf{h}^{(l+1)} = \text{MLP}(\text{LN}(\tilde{\mathbf{h}}^{(l)})) + \tilde{\mathbf{h}}^{(l)}
\end{equation}
where $\text{LN}$ is Layer Normalization, MSA is multi-head self-attention, and MLP is a position-wise feed-forward network with GELU activation.
{The MSA uses a causal attention mask, so that the hidden states $\mathbf{h}_i^{(l)}$ at position $i$ depends only on the preceding tokens $\mathbf{s}_{1:i}$. The final-layer hidden state $\mathbf{h}_i^{(L)} \in \mathbb{R}^d$ then serves as the shared input to the next-token prediction head and the goal-prediction head described below.}

\subsubsection{Next-Token Prediction Head}


\begin{figure}[t!]
  \centering
  \includegraphics[width=\linewidth]{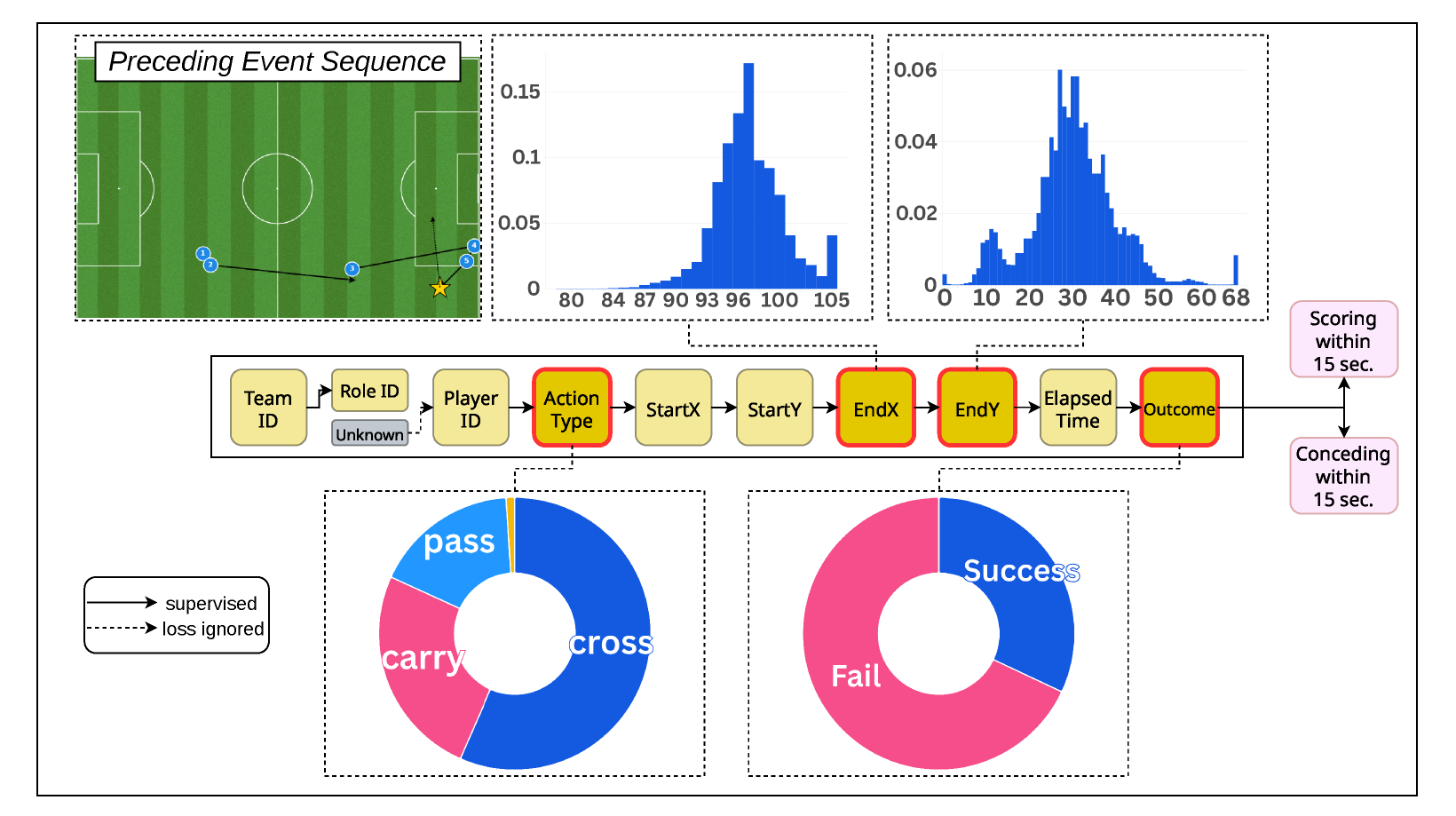}
  \caption{Next-token prediction over the preceding event sequence. The pitch shows the final five ball-progressing actions leading up to the target, which form the preceding event sequence that we serialize into tokens and feed to the model; the star ($\star$) marks the event the model predicts. At selected tokens we overlay the model's predicted distribution, shown as pie charts for action type and outcome and histograms for the end-location coordinates $x_t^{\text{end}}$ and $y_t^{\text{end}}$.}
  \label{fig:next_pred}
\end{figure}



The next-token prediction head applies a linear map $W_{\rm next} \in \mathbb{R}^{|\mathcal{V}| \times d}$ followed by a softmax to the final-layer hidden state $\mathbf{h}_i^{(L)}$, yielding a probability distribution over the entire vocabulary $\mathcal{V}$:
\begin{equation}
    \hat{\mathbf{y}}_{i+1} = (\hat{y}_{i+1,1}, \ldots, \hat{y}_{i+1,|\mathcal{V}|}) = \mathrm{softmax}\!\left( W_{\rm next} \mathbf{h}_i^{(L)} \right) \in [0,1]^{|\mathcal{V}|},
    \label{eq:next_head}
\end{equation}
where $\hat{y}_{i+1,s} = P(s \mid \mathbf{s}_{1:i}; \theta)$ for each $s \in \mathcal{V}$ estimates the probability of the next token being $s$. This makes the model predicts the fields of each event in the order listed in Table~\ref{tab:input_schema} conditioned on the preceding tokens. Fig.~\ref{fig:next_pred} illustrates this procedure with the estimated probabilities. Note that the player-ID field is excluded from prediction and instead resolved from the predicted team and role at inference time (Section~\ref{subsec:inference}), since training on it shapes the player embeddings more by team identity then by actual playing style (Section~\ref{sec:retrieval}).


\subsubsection{Goal Prediction Head}

At the last-token position $i(t) = 54 + 10t$ of each event $\mathbf{e}_t$, we attach a separate classification head that jointly estimates the goal-scoring probability $\hat{g}_{t}^{+} = P(g_t^+=1 \mid \mathbf{s}_{1:i(t)}; \theta)$ and the goal-conceding probability $\hat{g}_{t}^{-} = P(g_t^-=1 \mid \mathbf{s}_{1:i(t)}; \theta)$ within 15 seconds after $\mathbf{e}_t$. Specifically, we apply a linear map $W_{\rm goal} \in \mathbb{R}^{2 \times d}$ to the final-layer hidden state $\mathbf{h}_{i(t)}^{(L)}$, and a component-wise sigmoid $\sigma$ produces two probabilities:
\begin{equation}
    (\hat{g}_{t}^{+}, \hat{g}_{t}^{-}) = \sigma \left( W_{\rm goal} \mathbf{h}_{i(t)}^{(L)} \right) \in [0,1]^2.
    \label{eq:goal_head}
\end{equation}


\subsection{Multi-Task Training Objective}
\label{subsec:objective}
We train ScoutGPT with a composite loss that balances the two task capabilities:
\begin{equation}
      \mathcal{L} = \mathcal{L}_{\rm next} + \mathcal{L}_{\rm goal},
\end{equation}
where the next-token prediction loss $\mathcal{L}_{\rm next}$ and the goal prediction loss $\mathcal{L}_{\rm goal}$ are defined below.

The primary term is the next-token cross-entropy loss $\mathcal{L}_{\rm next}$, the negative log-likelihood of the true token sequence under the model:
\begin{equation}
    \small
    \mathcal{L}_{\rm next}
    = -\frac{1}{|\mathcal{D}|} \sum_{\mathbf{s} \in \mathcal{D}} \sum_{i \in \mathcal{I}_{\rm next}(\mathbf{s})} \log P(s_{i+1} \mid \mathbf{s}_{1:i}; \theta)
    = -\frac{1}{|\mathcal{D}|} \sum_{\mathbf{s} \in \mathcal{D}} \sum_{i \in \mathcal{I}_{\rm next}(\mathbf{s})} \log \hat{y}_{i+1,s_{i+1}}
\end{equation}
where $\mathcal{D}$ is the training dataset, $\mathcal{I}_{\rm next}(\mathbf{s})$ denotes the set of token positions at which a next token is predicted, and $\hat{y}_{i+1,s_{i+1}}$ is the probability that model assigns to the ground-truth token $s_{i+1}$ (Eq.~\ref{eq:next_head}). Minimizing $\mathcal{L}_{\rm next}$ maximizes the likelihood assigned to the observed events, training ScoutGPT to reproduce the game dynamics.

As motivated in Section~\ref{subsec:architecture}, player-ID prediction is excluded from thie objective. Since a token at position $i+1$ is predicted from position $i$, and the role field precedes the player-ID field in each event tuple (Table~\ref{tab:input_schema}), excluding the player-ID targets amounts to dropping every role-token position from $\mathcal{I}_{\rm next}(\mathbf{s})$:
\begin{equation}
    \mathcal{I}_{\rm next}(\mathbf{s}) = \mathcal{I}_{\rm event}(\mathbf{s}) - \mathcal{I}_{\rm role}(\mathbf{s})
\end{equation}
where $\mathcal{I}_{\rm event}(\mathbf{s})$ and $\mathcal{I}_{\rm role}(\mathbf{s})$ are the sets of event-token positions and role-token positions in $\mathbf{s}$, respectively.


In addition to next-token prediction, the goal prediction loss $\mathcal{L}_{\rm goal}$ is computed by applying binary cross-entropy (BCE) to the goal-scoring or goal-conceding predictions at the last token of each event. For each event $\mathbf{e}_t$, the predicted probabilities $\hat{g}_{t}^{+}$ and $\hat{g}_{t}^{-}$ from Eq.~\ref{eq:goal_head} are compared against the ground-truth labels $g_t^+, g_t^- \in \{0,1\}$ retrieved from the raw data, yielding the goal prediction loss:
\begin{equation}
    \mathcal{L}_{\rm goal} = \frac{1}{|\mathcal{D}|} \sum_{\mathbf{s} \in \mathcal{D}}
    \sum_{t=1}^T
    \left[
      \text{BCE}(\hat{g}_{t}^{+}, g_t^+)
      +
      \text{BCE}(\hat{g}_{t}^{-}, g_t^-)
    \right],
\end{equation}
where $\text{BCE}(\hat{g}, g) = -[g\log\hat{g} + (1-g)\log(1-\hat{g})]$.



\subsection{Inference with Structural Constraints}
\label{subsec:inference}
Generating realistic football sequences requires strict logical consistency between consecutive events. Standard sampling can produce syntactically valid but physically invalid sequences, so we mask out implausible tokens in terms of the past token sequence and resolve the acting player through a spatial heuristic.

\subsubsection{Invalid Token Masking}
To prevent the model from generating tokens that violate the game's logical constraints, we restrict the output distribution at each decoding step to the set of valid sequences. Concretely, at each position $i$, we apply a validity mask $\mathbf{m}_i = (m_{i,1}, \ldots, m_{i,|\mathcal{V}|})$ to the output logits $\mathbf{z}_i = (z_{i,1}, \ldots, z_{i,|\mathcal{V}|})$ based on the past token sequence $\mathbf{s}_{1:i}$:
\begin{equation}
    P(\cdot \mid \mathbf{s}_{1:i}) = \text{softmax}(\mathbf{z}_i + \mathbf{m}_i),
\end{equation}
where $\mathcal{V}$ is the token vocabulary, $m_{i,j} = 0$ if the $j$-th token in $\mathcal{V}$ is valid under $\mathbf{s}_{1:i}$ and $m_{i,j} = -\infty$ otherwise. Adding $-\infty$ drives the probability of invalid tokens to zero after the softmax, so only admissible tokens can be sampled.


\subsubsection{Spatial-Aware Entity Resolution}
After a team token $\hat{h}_t$ is generated, the model first predicts a compatible role token $\hat{r}_t$. The pair $(\hat{h}_t, \hat{r}_t)$ then narrows the candidate player tokens to those in the lineup whose team and role are $\hat{h}_t$ and $\hat{r}_t$, respectively. When there is only one candidate player, the corresponding player token is determined immediately. When multiple players share the same $(\hat{h}_t, \hat{r}_t)$ pair, the model selects the candidate whose reference location $\mu_p$ is closest to the most recent ball location $(\hat{x}_t,
\hat{y}_t)$, i.e., the end coordinates of the preceding event:
\begin{equation}
    p^* = \operatorname*{argmin}_{p \in \mathcal{P}(\hat{h}_t, \hat{r}_t)} \left\| (\hat{x}_t, \hat{y}_t) - \mu_p \right\|_2,
\end{equation}
where $\mathcal{P}(\hat{h}_t, \hat{r}_t)$ denotes the set of candidate players matching the generated team $\hat{h}_t$ and role $\hat{r}_t$. 
We use the preceding event's end location because the acting player is resolved before the current event's coordinates are generated, and in
practice it nearly coincides with the start location of the current event. The reference location $\mu_p$ of a known player (i.e., a player
observed in the training set) is set to the average location of their training-set events, which prevents leakage from evaluation matches. For
an unknown player, it falls back to a default location associated with their role.
Furthermore, the generator applies an ownership-lock mechanism tied to the VERSA possession state. While the validator indicates that a player
retains the ball, the lock fixes that player across consecutive in-possession events instead of resolving a new actor at each step. When
possession ends, the lock is released and the player is barred from immediate reselection for one event, stabilizing local event continuity.

\subsubsection{Dynamic Episode Termination}
Episode generation uses semantic stopping rules in addition to the generic EOS token. Decoding stops when (1) an EOS token is produced; (2) a set-piece restart action is generated—\textit{Corner}, \textit{Throw-in}, \textit{Free Kick}, or \textit{Goal Kick}; (3) a successful \textit{Shot} or \textit{Penalty Kick} results in a goal; or (4) \textit{Foul} or \textit{Own Goal}, is generated. These rules prevent generation beyond the logical boundary of the current phase of play. The complete action type taxonomy is listed in the Supplementary Material.

\section{Experiments}

We evaluate ScoutGPT on event data collected from South Korean K League across three axes: next-event prediction accuracy, goal prediction quality, and counterfactual transfer simulation. We first describe the experimental setup, then report main results and ablations, and finally assess simulation fidelity via self-to-self reconstruction.

\subsection{Experimental Setup}

\subsubsection{Dataset}
\begin{table}[t]
\centering
\caption{Summary of the dataset}
\label{tab:dataset_stats}
\begin{tabular}{llccccc}
\toprule
Split & Seasons & Matches & Episodes & Events & Events/ep & Players \\
\midrule
Train & 2021--2023 & 1{,}320 & 132{,}315 & 3{,}528{,}635 & 26.67 & 1{,}090 \\
Valid & 2024 & 462 & 43{,}579 & 1{,}277{,}169 & 29.31 & 848 \\
Test & 2025 & 501 & 47{,}046 & 1{,}324{,}363 & 28.15 & 859 \\
\bottomrule
\end{tabular}
\end{table}
We evaluate our model using event data from five seasons of the South Korean K League 1 and 2, spanning from $2021$ to $2025$. We standardize
all data according to the VERSA event representation~\cite{Jo2026VERSA}, a state-transition based verifier that defines the taxonomy of 29 action types and enforces football's logical constraints, correcting anomalies such as missing \textit{Pass Received} events or temporally inconsistent orderings. We partition the dataset chronologically for model training and evaluation. We use data from the $2021$, $2022$, and $2023$ seasons as the training set and the $2024$ season as the validation set, and reserve the $2025$ season for final testing. Detailed dataset statistics can be found in Table~\ref{tab:dataset_stats}.

\subsubsection{Baselines}

We evaluate ScoutGPT on two tasks, each with its own baselines. First, we compare against next-token predictors derived from prior work~\cite{MendesNeves2026Scalable}, namely an LSTM~\cite{hochreiter1997lstm}, a gradient-boosted CatBoost~\cite{Prokhorenkova2018CatBoost}, the MLP-based Large Events Model (LEM-MLP)~\cite{MendesNeves2024Towards}, and a Transformer variant (LEM-Transformer). The LEM Transformer adapts the Transformer backbone of NMSTPP~\cite{Yeung2025Transformer} to our output formulation, keeping its architecture but replacing the prediction heads with ones that match our multi-attribute event representation. We adapt these models rather than reuse them directly because existing sequence predictors such as Seq2Event~\cite{Simpson2022Seq2Event} and NMSTPP predict only coarse-grained action types (e.g., pass, dribble, shot), do not model additional event attributes such as player identity, spatial coordinates, temporal intervals, or action value, and do not incorporate the player conditioning that is central to our transfer-fit evaluation.

In addition, we compare ScoutGPT's goal prediction performance against the CatBoost
baseline~\cite{Decroos2020VAEP} and three neural sequence models, an LSTM~\cite{hochreiter1997lstm}, a GRU~\cite{cho2014gru}, and a
Transformer~\cite{Vaswani2017Attention}, each trained to predict the same goal labels from event context.



\subsection{Main Results}

\begin{table}[ht!]
\centering
\footnotesize
\setlength{\tabcolsep}{3.5pt}
\caption{Performance comparison across categorical and continuous attributes. For categorical attributes, Accuracy / F1 are reported; for continuous attributes, $R^2$ / MAE are reported.}
\label{tab:main_results}
\begin{tabular}{llcccc>{\columncolor{oursbg}}c}
\toprule
\textbf{Group} & \textbf{Attr.} & LSTM & CatBoost & \makecell{LEM-\\MLP} & \makecell{LEM-\\Transformer} & \textbf{ScoutGPT} \\
\midrule
\multirow{4}{*}{Cat.}
& Team  & 0.94 / 0.94 & 0.94 / 0.94 & 0.66 / 0.37 & \bftab{0.96} / 0.49 & 0.92 / \bftab{0.92} \\
& Role  & 0.77 / \bftab{0.67} & 0.76 / 0.66 & 0.71 / 0.32 & 0.45 / 0.22 & 0.63 / 0.47 \\
& Type  & 0.76 / 0.34 & 0.76 / 0.33 & 0.71 / 0.20 & 0.77 / 0.43 & \bftab{0.78} / \bftab{0.53} \\
& Out.  & 0.89 / 0.59 & 0.89 / 0.59 & 0.92 / 0.14 & 0.89 / 0.61 & \bftab{0.95} / \bftab{0.86} \\
\midrule
\multirow{5}{*}{Cont.}
& Start $x$ & 0.74 / 2.78 & 0.80 / 3.88 & 0.91 / 2.14 & 0.78 / 4.59 & \bftab{0.96} / \bftab{0.97} \\
& Start $y$ & 0.76 / 2.34 & 0.80 / 3.32 & \bftab{0.94} / 1.63 & 0.78 / 4.09 & 0.93 / \bftab{1.00} \\
& End $x$ & 0.61 / 7.26 & 0.70 / 7.55 & 0.48 / 6.27 & 0.67 / 8.12 & \bftab{0.89} / \bftab{4.11} \\
& End $y$ & 0.54 / 6.95 & 0.67 / 6.87 & 0.76 / 5.16 & 0.66 / 7.18 & \bftab{0.82} / \bftab{3.85} \\
& Time     & 0.38 / 0.53 & 0.49 / 0.64 & 0.46 / 1.03 & 0.54 / 1.42 & \bftab{0.71} / \bftab{0.75} \\
\bottomrule
\end{tabular}
\end{table}
\begin{table}[ht!]
\centering
\caption{Performance comparison on goal-scoring (GS) and goal-conceding (GC) prediction within 15 seconds.}
\label{tab:vaep_comparison}
\begin{tabular}{p{2.5cm} ccc ccc}
\toprule
& \multicolumn{3}{c}{\textbf{Goal-scoring (GS)}} & \multicolumn{3}{c}{\textbf{Goal-conceding (GC)}} \\
\cmidrule(lr){2-4} \cmidrule(lr){5-7}
\textbf{Method} & \textbf{AUC} & \textbf{Brier} & \textbf{ECE} & \textbf{AUC} & \textbf{Brier} & \textbf{ECE} \\
\midrule
CatBoost~\cite{Decroos2020VAEP} & \bftab{0.8424} & 0.0075 & \bftab{0.0003} & 0.8051 & 0.0021 & 0.00082 \\
LSTM~\cite{hochreiter1997lstm} & 0.8294 & 0.0230 & 0.0660 & 0.8080 & 0.0029 & 0.01383 \\
GRU~\cite{cho2014gru} & 0.8250 & 0.0205 & 0.0576 & 0.8091 & 0.0029 & 0.01376 \\
Transformer~\cite{Vaswani2017Attention} & 0.8277 & 0.0210 & 0.0589 & 0.7948 & 0.0031 & 0.01519 \\ \midrule\rowcolor{oursbg}
\textbf{ScoutGPT} & 0.8344 & \bftab{0.0069} & 0.0024 & \bftab{0.8153} & \bftab{0.0016} & \bftab{0.00081} \\
\bottomrule
\end{tabular}
\end{table}


Table~\ref{tab:main_results} reports end-to-end event modeling across event
formats and model families. ScoutGPT is strongest on structurally important
targets---type (0.78 / 0.53), outcome (0.95 / 0.86), end coordinates
(0.89 / 4.11 and 0.82 / 3.85), and time (0.71 / 0.75)---and also attains the best
start-$x$ (0.96 / 0.97). Gains over LEM Transformer are especially pronounced on
continuous variables (start-$x$ MAE 4.59 $\rightarrow$ 0.97, time MAE
1.42 $\rightarrow$ 0.75), indicating better spatial-temporal fidelity during
rollout. Although other models stay competitive on a few metrics (e.g., team/role
accuracy, start-$y$ $R^2$), ScoutGPT offers the most reliable overall trade-off
for realistic sequence continuation.


We then evaluate short-horizon goal prediction within 15 seconds using AUC, Brier
score, and ECE (Table~\ref{tab:vaep_comparison}). The goal-prediction head is
better calibrated for GC risk signals (higher GC AUC, lower calibration error),
while CatBoost gives stronger GS discrimination (higher GS AUC, lower ECE).
Crucially, because this head shares the backbone used for next-token prediction,
a single ScoutGPT model produces both the event predictions and the goal
probabilities, unifying the two separate baseline stages of event modeling and
value estimation. This supports coupling autoregressive sequence modeling with
explicit value supervision to improve goal-related signal quality beyond pure
token prediction.
\subsection{Ablation Study}

\begin{table}[t]
\centering
\caption{Ablation on lineup information and the context block. The Ours column is
the full model's absolute performance; the other columns report the change after
removing each component (Accuracy / F1 for categorical attributes, $R^2$ / MAE
for continuous). Improvement corresponds to positive changes for Accuracy, F1,
and $R^2$, and to negative changes for MAE; green and red mark improvement and
degradation.}
\label{tab:ablation}
\setlength{\tabcolsep}{3.5pt}
\begin{tabular}{ll>{\columncolor{oursbg}}ccccc}
\toprule
\textbf{Group} & \textbf{Attr.} & \textbf{Ours} & \textbf{w/o Context} & \textbf{w/o Lineup} & \textbf{w/o Both} \\
\midrule
\multirow{4}{*}{Cat.}
& Team
& 0.921 / 0.921
& \worse{-0.003} / \worse{-0.002}
& \worse{-0.008} / \worse{-0.007}
& \worse{-0.003} / \worse{-0.003} \\
& Role
& 0.634 / 0.469
& \improve{+0.002} / \improve{+0.047}
& \worse{-0.090} / \worse{-0.156}
& \worse{-0.069} / \worse{-0.128} \\
& Type
& 0.782 / 0.529
& \improve{+0.001} / \worse{-0.016}
& \worse{-0.002} / \worse{-0.006}
& \worse{-0.004} / \worse{-0.033} \\
& Out.
& 0.947 / 0.862
& \worse{-0.001} / \same{+0.000}
& \worse{-0.001} / \worse{-0.001}
& \same{+0.000} / \improve{+0.004} \\
\midrule
\multirow{5}{*}{Cont.}
& Start $x$
& 0.961 / 0.973
& \worse{-0.005} / \worse{+0.048}
& \improve{+0.004} / \improve{-0.020}
& \improve{+0.009} / \improve{-0.077} \\
& Start $y$
& 0.925 / 1.001
& \worse{-0.018} / \worse{+0.108}
& \worse{-0.029} / \worse{+0.164}
& \worse{-0.003} / \worse{+0.011} \\
& End $x$
& 0.886 / 4.105
& \worse{-0.011} / \worse{+0.143}
& \worse{-0.007} / \worse{+0.094}
& \worse{-0.012} / \worse{+0.169} \\
& End $y$
& 0.818 / 3.850
& \worse{-0.017} / \worse{+0.158}
& \worse{-0.015} / \worse{+0.117}
& \worse{-0.009} / \worse{+0.068} \\
& Time
& 0.711 / 0.751
& \worse{-0.184} / \worse{+0.069}
& \improve{+0.006} / \improve{-0.008}
& \worse{-0.212} / \worse{+0.097} \\
\bottomrule
\end{tabular}
\end{table}

To isolate what drives these gains, Table~\ref{tab:ablation} compares ablations on lineup and context usage. Removing lineup/context can improve a few isolated metrics (e.g., Role or Time), but the full model remains strongest on most high-impact targets, particularly Team consistency, Type F1, and spatial end-point quality (End $x$, End $y$). This pattern suggests that lineup-aware context is most beneficial for preserving coherent tactical structure, even when simplified variants can fit specific marginals slightly better.


\begin{table}[ht!]
\centering
\setlength{\tabcolsep}{4pt}
\caption{Average VAEP under different match minute and score-state conditions. 
The control value corresponds to the fully controlled reference condition and is fixed at 0.02678 for all rows. 
$\Delta$ denotes the change relative to this reference ($\Delta=\text{Simulated}-\text{Original}$).}
\label{tab:vaep_context}
\begin{tabular}{llccc}
\toprule
\textbf{Minute} & \textbf{Score state} & \textbf{Original} & \textbf{Simulated} & \textbf{$\Delta$} \\
\midrule
\multirow{3}{*}{0}
& Drawing  & 0.02678 & 0.023477 & $-$0.003303 \\
& Trailing & 0.02678 & 0.025916 & $-$0.000864 \\
& Leading  & 0.02678 & 0.026392 & $-$0.000388 \\
\midrule
\multirow{3}{*}{40}
& Drawing  & 0.02678 & 0.029647 & $+$0.002867 \\
& Trailing & 0.02678 & 0.029864 & $+$0.003084 \\
& Leading  & 0.02678 & 0.029644 & $+$0.002864 \\
\bottomrule
\end{tabular}
\end{table}

To further examine whether ScoutGPT captures context-dependent player impact under controlled perturbations, we conduct a context intervention
ablation on episodes of Jeonbuk Hyundai FC, reporting aggregated team-level episode VAEP across combinations of match minute and score state
(Table~\ref{tab:vaep_context}). At minute 0 the simulated play yields lower VAEP across all score states, most strongly when drawing
($\Delta=-0.0033$), reflecting a generally cautious style early in the second half. By minute 40 the effect reverses to higher VAEP across all
states, strongest when trailing ($\Delta=+0.0031$), reflecting a shift toward more proactive play, consistent with teams being most aggressive
in pursuit of an equalizer. Overall, ScoutGPT captures both context-dependent adaptation over time and meaningful differences in strategic
behavior across score states.
  
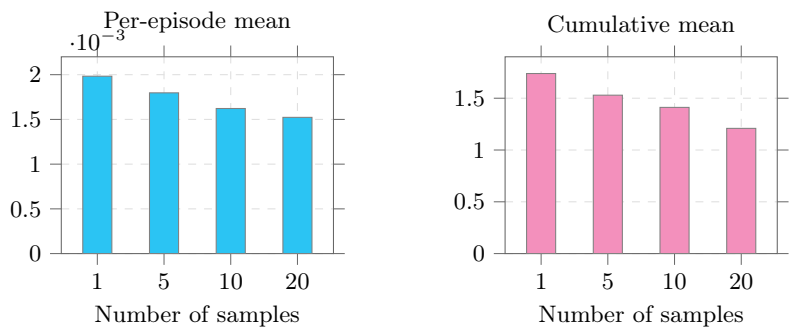
\begin{figure}[t!]
    \centering
    \begin{tikzpicture}

\begin{axis}[
    width=5.2cm,
    height=4.2cm,
    at={(0cm,0cm)},
    anchor=south west,
    ybar,
    bar width=11pt,
    ymin=0,
    ymax=0.0022,
    symbolic x coords={1,5,10,20},
    xtick=data,
    xlabel={Number of samples},
    title={Per-episode mean},
    scaled y ticks=true,
    yticklabel style={
        /pgf/number format/fixed,
        /pgf/number format/precision=3,
        font=\small
    },
    xticklabel style={font=\small},
    xlabel style={font=\small},
    title style={font=\small},
    enlarge x limits=0.18,
    tick align=outside,
    axis line style={black!60},
    tick style={black!60},
    grid=major,
    grid style={dashed, gray!25},
]

\addplot[
    draw=black!50,
    fill=cyan!65
] coordinates {
    (1, 0.001983)
    (5, 0.001797)
    (10,0.001622)
    (20,0.001523)
};

\end{axis}

\begin{axis}[
    width=5.2cm,
    height=4.2cm,
    at={(5.9cm,0cm)},
    anchor=south west,
    ybar,
    bar width=11pt,
    ymin=0,
    ymax=1.9,
    symbolic x coords={1,5,10,20},
    xtick=data,
    xlabel={Number of samples},
    title={Cumulative mean},
    yticklabel style={
        /pgf/number format/fixed,
        /pgf/number format/precision=2,
        font=\small
    },
    xticklabel style={font=\small},
    xlabel style={font=\small},
    title style={font=\small},
    enlarge x limits=0.18,
    tick align=outside,
    axis line style={black!60},
    tick style={black!60},
    grid=major,
    grid style={dashed, gray!25},
]

\addplot[
    draw=black!50,
    fill=magenta!55
] coordinates {
    (1, 1.738423)
    (5, 1.529375)
    (10,1.41149)
    (20,1.208931)
};

\end{axis}

\end{tikzpicture}
    \caption{Comparison of the mean absolute delta under different numbers of samples. The left shows the per-episode mean absolute delta, while the right shows the cumulative mean absolute delta.}
    \label{fig:sampling}
\end{figure}

Figure~\ref{fig:sampling} summarizes the discrepancy between ground-truth (GT) and self-to-self simulated episode VAEP across different numbers of samples. 
As the number of samples increases, both the per-episode mean and cumulative absolute differences consistently decrease. This indicates that larger sample sizes lead to more stable self-to-self simulation results and improve the agreement with the GT episode VAEP.

\begin{table}[ht!]
\centering
\footnotesize
\caption{Cross-season player retrieval performance using player embeddings learned from the 2021--2023 seasons. 
For each query embedding of a player from one season, retrieval is performed over embeddings from other seasons only. 
Evaluation is restricted to players who appear in at least two seasons. 
A query is counted as correct if an embedding of the same player from a different season is retrieved.}
\label{tab:cross_season_retrieval}
\begin{tabular}{lccc}
\toprule
Model & Top-1 (\%) & Top-5 (\%) & Top-10 (\%) \\
\midrule\rowcolor{oursbg}
ScoutGPT (Ours) & \bftab{9.20} & \bftab{21.97} & \bftab{30.90} \\\midrule
w/o position masking & 8.48 & 20.02 & 27.05 \\
Statistics-based embedding & 8.98 & 21.36 & 30.34 \\
\bottomrule
\end{tabular}
\end{table}
\section{Applications}

Beyond next-event prediction, ScoutGPT supports two applications that leverage the learned player representations and counterfactual generation capability: embedding-based player retrieval and hypothetical transfer simulation.

\subsection{Player Embedding for Similar Player Retrieval}
\label{sec:retrieval}
Table~\ref{tab:cross_season_retrieval} gives a quantitative view of the same
embedding space. Under \textit{role masking}, role tokens are hidden during
event prediction so the model must rely on player identity and event context
rather than explicit role labels. Same-player retrieval consistently improves
over both the no-masking variant and a stats-based embedding baseline. Masking
prevents role from acting as a shortcut bias, encouraging embeddings to encode
player-specific behavior.

\begin{figure}[h!]
    \centering
    \includegraphics[width=0.72\linewidth]{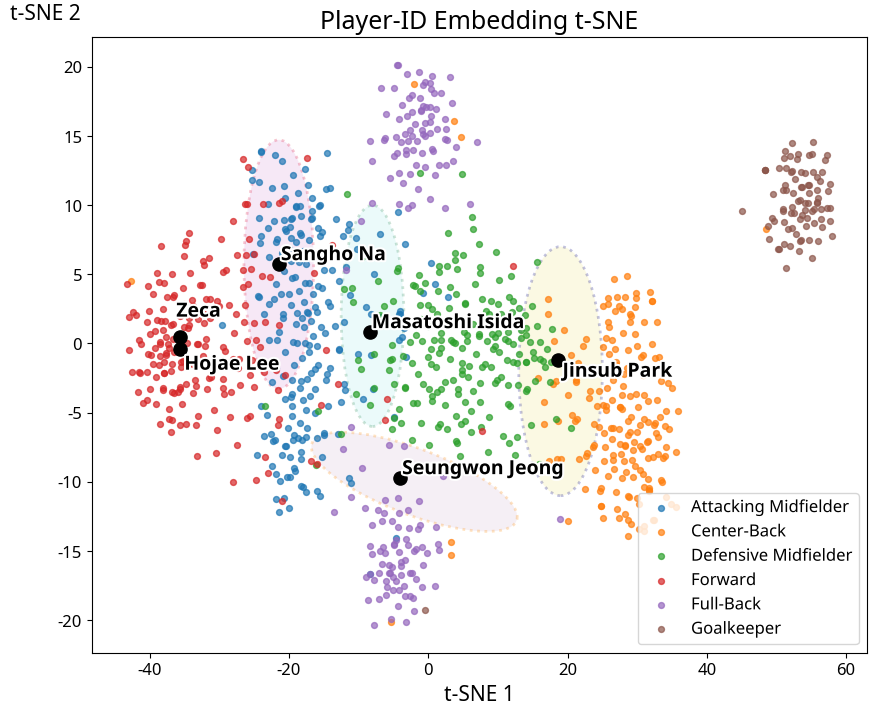}
    \caption{$t$-SNE projection of ScoutGPT player embeddings from the 2024 \textit{K League} season, colored by role.}
    \label{fig:player_embedding}
\end{figure}

To assess whether these embeddings capture role-related structure, we project
them into two dimensions with $t$-Distributed Stochastic Neighbor Embedding
($t$-SNE), shown in Figure~\ref{fig:player_embedding}, where each point is a
player colored by coarse role category.


Even with role masked, players separate by role, and the cluster hierarchy aligns with the spatial intuition of real football. Defensive midfielders are positioned between the center back cluster on the right and the attacking midfielder cluster on the left. Fullback clusters appear separated vertically based on their tactical roles on the field.  For instance, Jinsub Park exists between the defensive midfielder and center back clusters. Notably, these embedding patterns align well with their actual playing profiles, as all four players are known for their tactical versatility and have performed across multiple roles in real matches. A complete list of these multi-role players is provided in the Supplementary Material.




\subsection{Hypothetical Transfer Simulation}
Table~\ref{tab:vaep_k1_k2} evaluates transfer-fit prediction by comparing simulated post-transfer episode VAEP against ground truth and a naive baseline. Here, the naive projection estimates a player's next-season performance by simply extrapolating from the previous season's VAEP while adjusting only for playing time, without accounting for changes in tactical context, team structure, or role. Across all 40 transferred players, ScoutGPT reduces the mean absolute error from 1.84 (naive) to 1.25 (simulated), corresponding to a substantial relative error reduction. This indicates that the model captures context-dependent changes in player contribution more reliably than static carry-over assumptions. The representative examples also show that improvements are not confined to one role: gains appear for full-backs, wingers, and central midfielders, suggesting that the framework generalizes across distinct tactical functions.

\begin{table}[ht]
\centering
\caption{Comparison of post-transfer episode VAEP in 2025 using naive and simulated estimates against ground truth (GT). The top 40 transferred players are selected by post-transfer minutes and used to compute the overall average. Only five players are shown for readability. 
Columns ``Naive'', ``GT'', and ``Sim'' report episode VAEP sums, while |GT $-$ Sim| and |GT $-$ Naive| report absolute errors with respect to GT.}
\label{tab:vaep_k1_k2}
\begin{tabular}{llccccc}
\toprule
\textbf{Role} & \textbf{Player} & \multicolumn{3}{c}{\textbf{Episode VAEP Sum}} & \multicolumn{2}{c}{\textbf{Absolute Error w.r.t. GT}} \\
\cmidrule(lr){3-5} \cmidrule(lr){6-7}
 &  & \textbf{Naive} & \textbf{GT} & \textbf{Sim} & \textbf{|GT $-$ Sim|} & \textbf{|GT $-$ Naive|} \\
\midrule
\multicolumn{2}{c}{Average} & 4.85 & 4.71 & 4.59 & \bftab{1.25} & 1.84 \\
\midrule
Left Back & Jinsu Kim & 7.00 & 11.07 & 11.63 & \bftab{0.56} & 4.07 \\
Left Wing & Reis & 16.15 & 12.19 & 12.06 & \bftab{0.13} & 3.96 \\
Center Back & Hoik Jang & 3.79 & 4.78 & 5.83 & 1.05 & \bftab{0.99} \\
Central Midfielder & Jihoon Cho & 8.83 & 5.09 & 5.31 & \bftab{0.22} & 3.74 \\
Left Back & Juyong Lee & 4.61 & 6.51 & 7.31 & \bftab{0.80} & 1.90 \\
\bottomrule
\end{tabular}
\end{table}
Jinsu Kim provides a clear example of the practical value of context-aware simulation. The naive estimate (7.00) substantially underestimates his observed post-transfer contribution (GT sum = 11.07), whereas ScoutGPT predicts 11.63, yielding a much smaller error (0.56 vs. 4.07). This aligns with real-world outcomes: his move was widely regarded as highly successful relative to initial expectations, and he was appointed team captain in the following season. This case highlights how counterfactual sequence modeling can identify upside that is missed by static baseline forecasts.


\section{Conclusion}

We present ScoutGPT, a player-conditioned, value-aware autoregressive framework
for football event modeling. By jointly predicting event attributes and residual
on-ball value, it captures both local action structure and downstream tactical
impact, outperforming sequence-based baselines in next-event prediction, spatial
precision, and future contribution estimation while learning interpretable player
embeddings without role supervision. Counterfactual player substitution further
enables transfer-fit evaluation under new tactical contexts. Future work will
integrate tracking-based trajectory signals for off-ball behavior and extend
generation to full-possession or match-level simulation.

\section*{Acknowledgements}

The authors gratefully acknowledge the K League Technical Study Group (TSG) for providing the K League match data used in this study. We also sincerely thank Bepro11 for their efforts in collecting and processing the event and tracking data. Miru Hong, Geonhee Jo, Hyeokje Cho, and Sang-Ki Ko were supported by the National Research Foundation of Korea (NRF) grant funded by the Korean government (MSIT)~(No. RS-2024-00456065).

\bibliographystyle{splncs04}
\bibliography{ref}
%

\FloatBarrier
\appendix
\setcounter{table}{0}
\setcounter{figure}{0}
\setcounter{equation}{0}
\renewcommand{\thetable}{S\arabic{table}}
\renewcommand{\thefigure}{S\arabic{figure}}
\renewcommand{\theequation}{S\arabic{equation}}
\section{Appendix}

\subsection{Episode Component}
Each episode is prefixed with a fixed context block that encodes the match setup before any event token is generated. Between the \texttt{<CTX>} and \texttt{</CTX>} delimiters, the block lists the home and away teams in order; for each team it places the team token ($u_{\text{H}}$ or $u_{\text{A}}$) followed by the (role, player) pairs $(r_i, p_i)$ of its eleven on-pitch players, and ends with the match-state vector $\mathbf{q}$ over period, minute, score, and cards:
\begin{equation}
\begin{aligned}
\texttt{<CTX>} \; 
&u_{\text{H}}\; r_1^{(\text{H})} p_1^{(\text{H})} \cdots r_{11}^{(\text{H})} p_{11}^{(\text{H})} \\   
&u_{\text{A}}\; r_1^{(\text{A})} p_1^{(\text{A})} \cdots r_{11}^{(\text{A})} p_{11}^{(\text{A})} \\
&\mathbf{q}\; \texttt{</CTX>}
\end{aligned}
\end{equation}
The entries between the \texttt{<CTX>} and \texttt{</CTX>} delimiters form the 54 context tokens $(c_1, \dots, c_{54})$ used throughout the main paper, where the match-state vector $\mathbf{q}$ expands into eight tokens (period, minute, home and away goals, and home and away yellow and red cards). Table~\ref{appendix:event_types} lists the action-type taxonomy from which the event tokens are drawn. Table~\ref{tab:role_abbreviations} gives the full names of the positional role abbreviations used for $r_k$ and $r_t$.

\begin{table}[H]
    \centering
    \setlength{\tabcolsep}{6pt}
    \caption{Action-type taxonomy (29 types), grouped into six functional
    categories. \emph{Count} and \emph{\%} report the number and share of events
    of each type over the full dataset (6{,}130{,}167 events). Each event carries
    a binary outcome $a\in\{0,1\}$: $a=1$ when the action achieves its intended
    objective (e.g.\ a pass or cross reaching a teammate, a shot resulting in a
    goal, a take-on or tackle retaining or regaining possession), and $a=0$
    otherwise. Types whose occurrence already implies their result (e.g.\ Carry,
    Recovery, Interception) are always recorded as successful, whereas terminal
    negative events (e.g.\ Foul, Error, Own Goal) are always unsuccessful.}
    \label{appendix:event_types}
    \begin{tabular}{llrr}
    \toprule
    \textbf{Category} & \textbf{Action Type} & \textbf{Count} & \textbf{\%} \\
    \midrule
    \multirow{8}{*}{On-ball}
    & Pass & 1{,}901{,}605 & 31.02 \\
    & Pass Received & 1{,}742{,}075 & 28.42 \\
    & Carry & 841{,}879 & 13.73 \\
    & Recovery & 263{,}881 & 4.30 \\
    & Cross & 70{,}332 & 1.15 \\
    & Shot & 50{,}165 & 0.82 \\
    & Take-on & 26{,}338 & 0.43 \\
    & Penalty Kick & 648 & 0.01 \\
    \midrule
    \multirow{8}{*}{Defensive}
    & Interception & 300{,}758 & 4.91 \\
    & Duel & 187{,}487 & 3.06 \\
    & Clearance & 128{,}011 & 2.09 \\
    & Intervention & 102{,}492 & 1.67 \\
    & Tackle & 93{,}312 & 1.52 \\
    & Block & 71{,}685 & 1.17 \\
    & Aerial Clearance & 6{,}237 & 0.10 \\
    & Defensive Line Support & 3{,}111 & 0.05 \\
    \midrule
    \multirow{3}{*}{Goalkeeping}
    & Catch & 8{,}920 & 0.15 \\
    & Hit & 8{,}156 & 0.13 \\
    & Parry & 6{,}631 & 0.11 \\
    \midrule
    \multirow{4}{*}{Set-piece}
    & Throw-in & 93{,}493 & 1.53 \\
    & Free Kick & 58{,}664 & 0.96 \\
    & Goal Kick & 35{,}303 & 0.58 \\
    & Corner & 19{,}944 & 0.33 \\
    \midrule
    \multirow{2}{*}{Episode-ending}
    & Foul & 55{,}458 & 0.90 \\
    & Own Goal & 141 & 0.00 \\
    \midrule
    \multirow{4}{*}{Other}
    & Error & 46{,}431 & 0.76 \\
    & Pause & 4{,}986 & 0.08 \\
    & Goal Miss & 1{,}225 & 0.02 \\
    & Goal Post & 799 & 0.01 \\
    \bottomrule
    \end{tabular}
\end{table}
\begin{table}[H]
\centering
\footnotesize
\caption{List of roles used as the role tokens $r_k$ (context) and $r_t$ (event).}
\label{tab:role_abbreviations}
\begin{tabular}{llll}
\toprule
Abbr. & Full name & Abbr. & Full name \\
\midrule
GK  & Goalkeeper                    & CAM & Central Attacking Midfielder \\
CB  & Center Back                   & LM  & Left Midfielder \\
LB  & Left Back                     & RM  & Right Midfielder \\
RB  & Right Back                    & LW  & Left Winger \\
LWB & Left Wing Back                & RW  & Right Winger \\
RWB & Right Wing Back               & CF  & Center Forward \\
CDM & Central Defensive Midfielder  & LF  & Left Forward \\
CM  & Central Midfielder            & RF  & Right Forward \\
\bottomrule
\end{tabular}
\end{table}
\subsection{Multi-Position Players}
Table~\ref{tab:multi_role_players} lists players who have appeared in multiple positional roles across the dataset. These players tend to lie at the boundaries between role clusters in the learned embedding space, as discussed in the player-embedding retrieval analysis of the main paper, consistent with their versatile real-world usage.
\begin{table}[H]
\centering
\footnotesize
\caption{Distribution of representative multi-role players. Minutes indicate total playing time in each role.}
\label{tab:multi_role_players}
\begin{tabular}{p{3cm}p{8cm}}
\toprule
\textbf{Player} & \textbf{Roles Played (minutes)} \\
\midrule
Jinsub Park & CB (4,383), CDM (2,081), CM (1,849) \\
Masatoshi Ishida & CM (2,282), CAM (1,693), RW (287), CF (237), RF (141), LM (90), LW (77), LF (67) \\
Sangho Na & LW (3,670), RW (1,508), LM (1,073), RM (692), CF (451), LF (405), RF (90) \\
Seungwon Jeong & RWB (2,355), CM (2,021), RW (450), RM (270), RB (199), CAM (90) \\
\bottomrule
\end{tabular}
\end{table}
\end{document}